\pdfoutput=1

\documentclass[11pt]{article}

\usepackage[]{ACL2023}

\usepackage{times}
\usepackage{latexsym}

\usepackage[T1]{fontenc}

\usepackage[utf8]{inputenc}

\usepackage{microtype}

\usepackage{inconsolata}
\usepackage{comment}

\usepackage{todonotes} 
\usepackage{framed}
\usepackage{caption}

\title{Dealing with Semantic Underspecification in Multimodal NLP}

\author{Sandro Pezzelle\\
  Institute for Logic, Language and Computation\\
  University of Amsterdam\\
  \texttt{s.pezzelle@uva.nl}} 

\begin{document}
\maketitle
\begin{abstract}
Intelligent systems that aim at mastering language as humans do must deal with its \textit{semantic underspecification}, namely, the possibility  
for a linguistic signal to convey 
only part of the information needed for communication to succeed.
Consider the usages of the pronoun \textit{they}, which can 
leave the gender and number of its referent(s) underspecified.
Semantic underspecification is not a bug but a crucial language feature that boosts its storage and processing efficiency. Indeed, human speakers can quickly and effortlessly integrate semantically-underspecified linguistic signals with a wide range of non-linguistic information, e.g., the multimodal context, social or cultural conventions, and shared knowledge. Standard
NLP models have, in principle, no or limited access to such extra information, while multimodal systems 
grounding language into other modalities, 
such as vision, are naturally equipped to account for this phenomenon.
However, we show that   
they struggle with it, which could negatively affect their performance and
lead to harmful consequences when used for applications. In this position paper, we argue that our community should 
be aware of semantic underspecification if it aims to develop language technology that can successfully interact with human users.  We discuss some applications where mastering it is crucial and outline a few directions toward achieving this goal.
\end{abstract}

\section{Introduction}\label{sec:intro}

\emph{They put the flowers there.}
Speakers of a language hear sentences like this every day and have no trouble understanding what they mean---and what message they convey. This is because, in a normal state of affairs, they can count on a wide range of information from the surrounding context, personal knowledge and experience, social or cultural conventions, and so on. Upon hearing this sentence, for example, they would know that flowers go into vases, look in the direction where their interlocutor nodded their chin, see 
a vase with tulips on the windowsill,
and \textit{infer} that 
this is where someone put the flowers.
Every time listeners need to count on extra, non-linguistic information to understand a linguistic signal, like in this example, it is because the language used is semantically \textit{underspecified}~\cite{ferreira2008ambiguity,frisson2009semantic,harrisCH1}.
In the example above, the locative adverb \emph{there} leaves underspecified a location---where the flowers were put---which would instead be explicitly provided in the  semantically more specified sentence \emph{They put the flowers in the light blue vase on the windowsill at the end of the hallway}.
According to linguists, indeed, adverbs of place (\textit{here}, \textit{there}) are typical examples of semantically underspecified words, as well as adverbs of time (\textit{now}, \textit{today}), demonstratives (\textit{this}, \textit{that}),
quantifiers (\textit{few}, \textit{many}), tensed expressions, and some usages of personal pronouns~\cite{lappin2000intensional,harrisCH1}.

The reason why semantic underspecification is so widespread has to do with language efficiency, which is a trade-off between informativeness and conciseness~\cite{zipf,goldberg2022good}.
Underspecified words can be used in many communicative occasions with varying meanings and intentions~\cite{harrisCH1}, 
which 
prevents speakers from 
fully articulating 
every nuance of meaning every time they talk~\cite{piantadosi2012communicative}. Indeed, planning and producing utterances---but also  speech~\cite[see][]{levinson2000presumptive}---is cognitively expensive~\cite{trott2022languages}.
The use of
underspecified language, at a first sight, seems to go against the view that language is a cooperative system~\cite{Grice75,tomasello2005constructing}
and can indeed explain cases where communication appears to be \emph{egocentric} rather than cooperative~\cite{keysar2007communication}. 
However, a wealth of studies has shown that humans are extremely good at making inferences~\cite{grice1969utterer,sedivy1999achieving} and that this ability is cognitively cheaper than speaker articulation, which is rather demanding and time-consuming~\cite{levinson2000presumptive}.
Upon hearing a semantically underspecified sentence,
human speakers can
quickly and effortlessly integrate linguistic and non-linguistic information~\cite{harrisCH1}. In this light,~\citet{levinson2000presumptive} proposed that semantic underspecification gives rise to processing efficiency besides boosting storage efficiency.

Semantic underspecification
allows our limited repertoire of symbols to be used in many contexts and with different intentions without compromising its communicative effectiveness. For example, we can use the pronoun \textit{they} to omit a person's gender or refer to a group of friends; the locative \textit{here} to refer to a free table at a caf{\'e} or the institution you work for.
Semantic underspecification 
is not a bug but a crucial feature of language that is ubiquitous in human communication~\cite{harrisCH2}. As such,
any intelligent system that aims at mastering language as humans do must be able to properly deal with it.
This is particularly the case for models of natural language understanding and generation that 
have access to non-linguistic information~\cite{bender2020climbing,bisk2020experience}, e.g., models integrating language and vision that account for 
the multimodality of language~\cite{harnad1990symbol}.
These models must be able to 
understand and generate sentences like \emph{They put the flowers there}, provided that a relevant visual context is present and there is a clear communicative goal.
This is a mandatory requirement if we want to  use these systems to model real communicative scenarios or embed them in applications that interact with human users.

In this position paper, we argue 
that semantic underspecification should be high on the NLP community agenda, particularly within approaches combining language and vision.
We report that SotA multimodal NLP models struggle with it, and advocate a comprehensive, thorough investigation of the phenomenon along several research directions and concrete steps.
Mastering semantic underspecification is a long-term goal that implies shifting the paradigm to a scenario where models use language as humans do, that is, \textit{with a communicative goal}.
In line with what was argued elsewhere~\cite{bisk2020experience,giulianelli2022towards,fried2022pragmatics}, we believe the time is ripe for such a change.

\section{How Do Multimodal Models Deal with Semantic Underspecification?}

The field of multimodal or visually grounded NLP is currently dominated by pre-trained multimodal Transformers. 
Since their introduction, models like CLIP~\cite{radford2021learning}, LXMERT~\cite{tan2019lxmert}, VisualBERT~\cite{li2019visualbert}, ViLBERT~\cite{lu2019vilbert}, VL-BERT~\cite{su2019vl}, UniT~\cite{hu2021unit}, VILLA~\cite{gan2020large}, UNITER~\cite{chen2020uniter}, VinVL~\cite{zhang2021vinvl}, ViLT~\cite{kim2021vilt}, and mPLUG~\cite{li2022mplug}, \textit{inter alia}, 
have rapidly become the new
state-of-the-art in virtually every language and vision task.
Among other tasks, these models achieve unprecedented performance in 
describing an image in natural language~\cite{lin2014microsoft}, finding the best image for a given language query~\cite{plummer2015flickr30k}, answering fine-grained questions about the content of an image~\cite{antol2015vqa,krishna2017visual,hudson2019gqa}, reasoning over objects and object relations~\cite{johnson2017clevr,suhr2019corpus}, and entertaining a visually-grounded dialogue by asking and answering questions~\cite{de2017guesswhat,das2017visual}.

These models differ from each other in several dimensions. For example, they either concatenate and jointly process the visual and textual embeddings (\textit{single-stream} models), or process the two modalities by means of separate encoders with an optional cross-modal fusion (\textit{dual-stream} models); or, they use visual features extracted with either CNN-based~\cite[e.g., region features from Faster R-CNN;][]{ren2015faster}
or Transformer-based~\cite[e.g., image features from Vision Transformer, ViT;][]{dosovitskiy2020image} image encoders. However, they share both the same underlying architecture, which is based on Transformers, and training regime, which leverages a massive amount of multimodal data and a few common learning objectives. One of the most popular learning objectives is Image-Text Matching (ITM), which
maximizes the similarity between an image and a language fragment that is well \textit{aligned} with it.
As a result of this training regime, these models are impressively good at judging whether a sentence is a good/bad (true/false) description of the content of an image. This is particularly the case for CLIP, which is optimized for the task and can almost perfectly spot word-level inconsistencies between an image and a sentence, as the ones included in the FOIL dataset by~\citet{shekhar2017foil}~\cite[results reported in][]{parcalabescu2022valse}.

Given this impressive performance, it is reasonable to expect that these models are robust to semantically underspecified language. Describing an image, asking a question, or entertaining a conversation about it are all communicative scenarios
that admit a varying degree of semantic underspecification.
For example, the question \textit{What are they doing?} referred to a visual context with people playing an unusual sport is perfectly acceptable---and indeed likely to be asked; or, the sentence \textit{A person is typing on their laptop} to describe an office environment is not only a very good description of that context but perhaps even a desirable one.
Therefore, mastering semantically underspecified language is a requisite for any multimodal NLP model which aims at both genuinely solving these tasks and being used for user-facing applications.

\begin{figure*}[t!]\small \centering
\fbox{
\begin{minipage}{0.25\textwidth}
\textsc{\textcolor{white}{TTT}}\\[6.5pt]
\includegraphics[width=\textwidth]{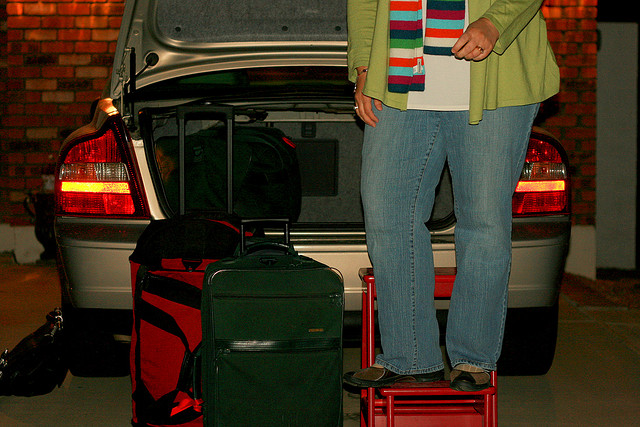} 
\end{minipage}
\hspace*{\fill}
\begin{minipage}{0.71\textwidth}
\begin{minipage}{0.2\textwidth}
\begin{tabular}{l@{\ }l}
\textsc{Type}\\[5pt]
\texttt{Original}\\[5pt]
\texttt{Quantity}\\
\texttt{Gender}\\
\texttt{Gender+Number}\\
\texttt{Location}\\
\texttt{Object}\\
\texttt{Full}\\ 
\end{tabular}
\end{minipage} 
\begin{minipage}{0.4\textwidth}
\begin{tabular}{l@{\ }l}
\textsc{Description}\\[5pt]
The woman is standing above the two packed suitcases.\\[5pt]
The woman is standing above \textbf{some} packed suitcases.\\
The \textbf{person} is standing above the two packed suitcases.\\
\textbf{They} are standing above the two packed suitcases.\\
The woman is standing \textbf{here}.\\
The woman is standing above \textbf{this}.\\
\textbf{They are doing something here.}\\
\end{tabular}
\end{minipage} 
\hspace*{\fill}
\begin{minipage}{0.1\textwidth}
\centering
\begin{tabular}{l@{\ }l}
\textsc{Score}\\[5pt]
0.8565\\[5pt]
0.8275\\
0.7608\\
0.7435\\
0.5537\\
0.4931\\
0.4646\\
\end{tabular}
\end{minipage}
\end{minipage}
}
\caption{
For one image, \texttt{COCO\_train2014\_000000394151.jpg}, we report one original description from COCO and the six corresponding semantically underspecified descriptions, along with their CLIPScore.}
\label{fig:extraction}
\end{figure*}

\subsection{Proofs of Concept}

To scratch the surface of the problem,
we carry out two Proofs of Concept (hence, PoCs) using image descriptions and the CLIP model. 
When talking about a visual context, speakers of a language can convey the same message with varying levels of semantic specification.
For example, they can describe someone waiting for the bus by referring to them as \textit{an elderly lady}, \textit{a woman}, \textit{a person}, or \textit{they}. Similarly, they can mention a location, i.e., \textit{the bus stop}, or use the locatives \textit{here} or \textit{there}; an object, i.e., \textit{the bus}, or use the demonstratives \textit{this} or \textit{that}; and so on.
This is possible because the visual context provides enough information for the addressee to understand the message, even when it is extremely semantically underspecified.

Almost by definition, standard image descriptions as those in COCO~\cite{lin2014microsoft} are semantically \textit{overspecified}. Indeed, they are meant to serve as a natural language `translation' of the content of an image, to make it available to someone who does not have access to the image~\cite[for a discussion on this point, see, e.g.,][]{kreiss2021concadia}. As such, these descriptions fully specify a wide range of semantic aspects that would be reasonably left underspecified if the visual context was available to both interlocutors.
As mentioned above, CLIP is extremely good at assessing whether a description is good for an image. 
As such, it is reasonable to expect that the model
should not be  
affected by the degree of semantic specification of the description, provided that it is valid for the image. To illustrate, a model should similarly score the descriptions \textit{A woman waiting for the bus} and \textit{A person waiting for the bus} in relation to the visual context described above. Moreover, a semantically valid underspecified description must always be better than an unrelated, overspecified description.

In the two PoCs below, we explore these two hypotheses. Note that we do so for illustrative purposes, highlighting general trends that can be useful for further, more thorough research. Moreover, it is worth stressing that, while we employ CLIP due to its effectiveness and accessibility, the point we make is more general in scope than focused on this specific model. The point is that \textit{models should not be affected by semantic underspecification when assessing the validity or applicability of an image description.} Concretely, we 
use 100 
images and corresponding descriptions (495 in total) from the 2014 train partition of COCO. Data and code available at: \url{https://github.com/sandropezzelle/sunglass}

\paragraph{Are Underspecified   Descriptions as Good as Overspecified Ones?}

In this PoC, we are interested to check whether CLIP is robust to semantic underspecification. The expectation is that the model should assign the same or a similar alignment score
to image descriptions with a varying level of semantic specification, provided that these descriptions are semantically correct for the image.

We compute CLIPScore for each of the 495 $\langle$image, description$\rangle$ pairs in our sample and select the 100 with the highest score. We refer to these 100 descriptions as \texttt{Original}. We then create up to 6 underspecified versions of each description in \texttt{Original} by manually perturbing their text to account for various underspecification phenomena. Such an annotation task was performed by a single annotator, the author of this paper, with a background in formal and computational linguistics.
Perturbations are carried out only where possible (thus, not all descriptions have all 6 versions), without altering the grammatical structure  of the sentence. The semantic underspecification phenomena we consider are illustrated in the example in Figure~\ref{fig:extraction} and described below:

\begin{itemize}
\item \texttt{Quantity:} We replace numbers (e.g., \textit{two}) and quantity expressions (e.g., \textit{a couple}) with the   quantifier \emph{some}  
\item \texttt{Gender:} We replace gender-marked (e.g., \textit{woman}) and age-marked (e.g., \textit{children}) nouns with the hypernyms \emph{person} or \emph{people}
\item \texttt{Gender+Number:} We replace any NPs in subject position, either singular or plural, with the pronoun \emph{they}, and harmonize verb agreement
\item \texttt{Location:} We replace PPs introduced by a preposition of place (e.g., \textit{at}) with the locatives \textit{here} or \textit{there}
\item \texttt{Object:} We replace NPs, typically in object position, with the demonstratives \textit{this} or \textit{that}
\item \texttt{Full:} We replace the entire sentence with the fully underspecified one \textit{They are doing something here.}  
\end{itemize} 

\begin{figure}[t!]
\centering
\includegraphics[width=\columnwidth]{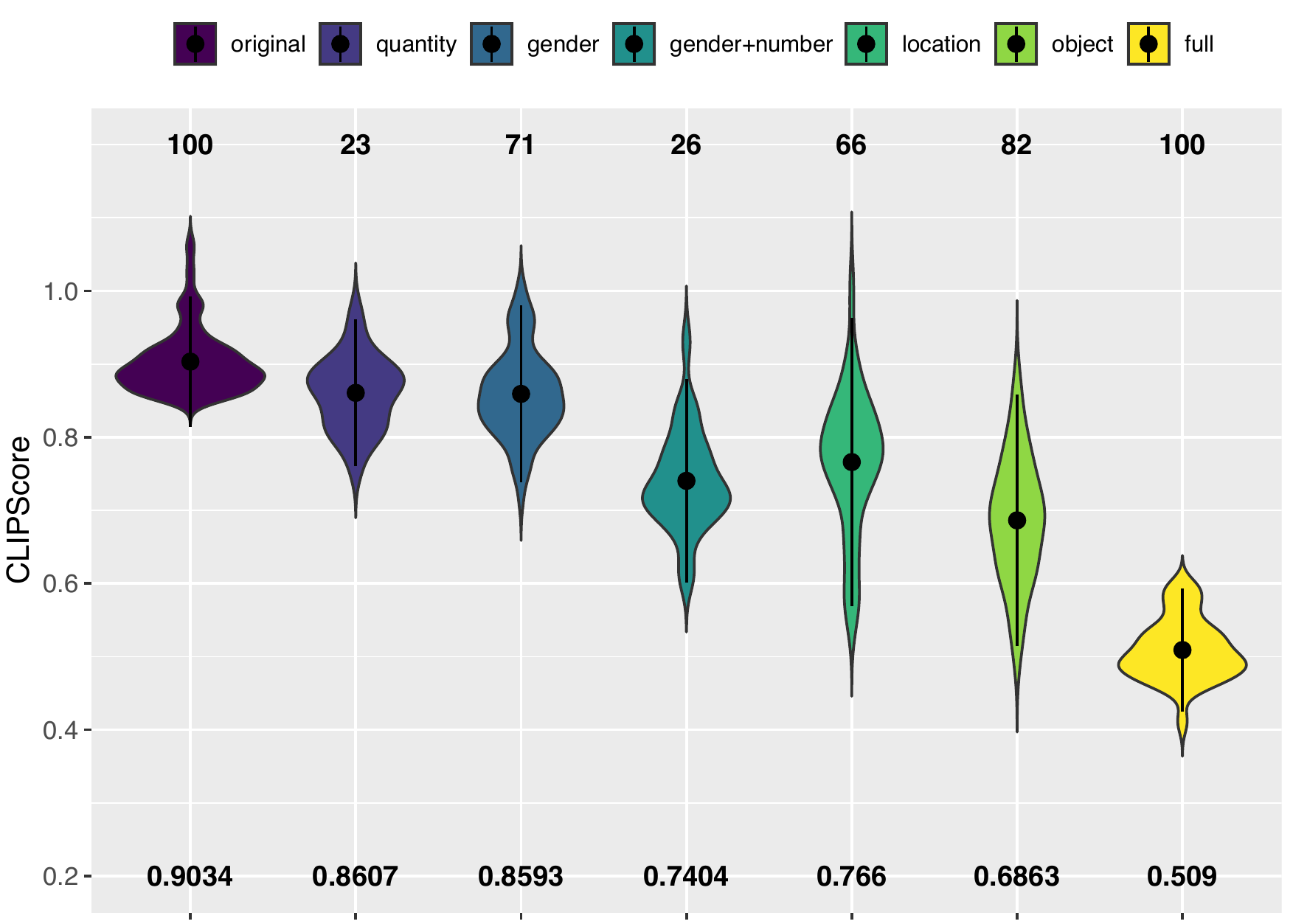}
\caption{Violin plot reporting the distribution of CLIPScore values per image description type. The dot and bars in each violin stand for the mean and standard deviation, respectively. The mean of each violin is also given at the bottom of the plot, while the integer at the top reports the sample size. Best viewed in color.}
\label{fig:violin}
\end{figure}

We compute CLIPScore for each underspecified description. In Figure~\ref{fig:violin}, we report the distribution of these scores against each phenomenon. As can be seen, underspecified descriptions achieve (much) lower scores compared to \texttt{Original} ones. For example, a perturbation as harmless as replacing the subject with the 
pronoun \emph{they} leads to a $\sim$16-point average decrease in CLIPScore, while the gap increases to $\sim$40 points when considering \texttt{Original} against the fully underspecified description \textit{They are doing something here.} All the scores for one specific example are reported in Figure~\ref{fig:extraction}.

These observations are surprising and go against our expectations that underspecified descriptions, if semantically valid, should be considered as good as overspecified ones. Indeed, \textit{why should a sentence containing a quantifier, a pronoun, or a locative be considered a poor description of a visual context?}
One possible explanation is that models like CLIP 
are sensitive to the amount of \textit{detail} provided by an image description. More specifically, the more words there are in the sentence with a clear and unique visual referent, the more the description is deemed `aligned' to an image. Using the terminology introduced by~\citet{kasai-etal-2022-transparent} to evaluate image captioning metrics, the model would be good at capturing an image description's \textit{recall}, i.e., the extent to which the salient objects in an image are covered in it; on the other hand, it would poorly capture a description's \textit{precision}, i.e., the degree to which it is precise or valid for a given image.\footnote{We thank the anonymous reviewer who referred us to this work.}
If this was the case, models like CLIP would end up always considering underspecified descriptions as worse than overspecified ones, which naturally raises questions about their robustness and applicability to a possibly wide range of scenarios.

\begin{figure}[t!]\small \centering
\fbox{ 
\begin{minipage}{0.95\linewidth}
\includegraphics[width=1\linewidth]{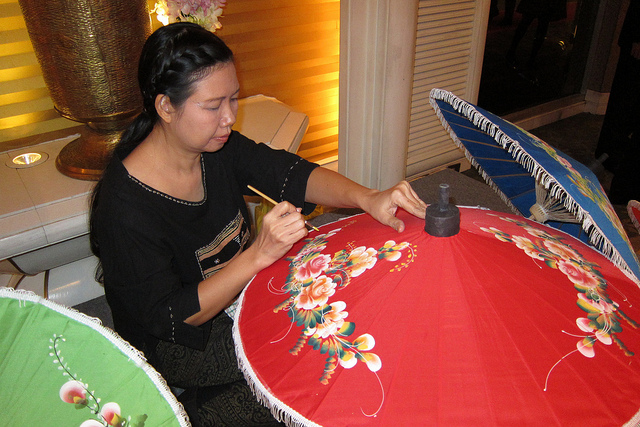}

\vspace{5pt}

\begin{minipage}{1\linewidth}
\begin{tabular} {@{}ll} 
\texttt{Full} They are doing something here.\\[5pt]
\texttt{r1} A woman in a white dress is sitting with her cell phone.\\
\texttt{r2} A girl with long brown hair with streaks of red lays\\\hspace{13pt}on a bed and looks at an open laptop computer.\\
\texttt{r3} A lady holding a bottle of ketchup and a dog in a hot\\\hspace{13pt}dog bun costume.\\
\texttt{r4} An infant sits next to a stuffed teddy bear toy.\\
\texttt{r5} Woman sitting on a bench holding a hotdog in her hand\\
\texttt{r6} Two small children playing with their refrigerator\\\hspace{13pt}magnets.\\
\end{tabular}
\end{minipage}
\end{minipage} 
}
\caption{
For \texttt{COCO\_train2014\_000000205931.jpg}, our fully underspecified image description achieves a lower CLIPScore than six randomly picked overspecified descriptions, although these are clearly wrong.
} \label{fig:errors}
\end{figure}

\paragraph{Are Underspecified Descriptions Better than Unrelated Ones?}

Even if CLIP was sensitive to the amount of detail provided by an image description (the more, the better), a valid underspecified description should always be deemed more aligned than an unrelated, overspecified one.
That is, even a highly underspecified sentence like
\emph{They are doing something here}---if semantically valid for the image, which is the case in our small sample---should always be preferred over a description that is fully unrelated to the image.
To test this hypothesis, we experiment with this \texttt{Full} description, and, for each image, we test it against 10 randomly sampled \texttt{Original} descriptions of other images.
Surprisingly, for 82 images out of 100, at least one random caption achieves a higher CLIPScore than \texttt{Full}. 
While the actual numbers may depend on the number and the type of random descriptions being sampled, some qualitative observations are helpful to highlight the behavior of the model.
Consider, as an example, the case reported in Figure~\ref{fig:errors}, where 6 out of 10 unrelated descriptions are deemed better than our fully underspecified one. By looking at these randomly picked sentences, it is apparent that none of them is a valid description of the image. At the same time, the model prefers them over a valid, though highly underspecified, description.

There are various possible explanations for this behavior. For example, the model could be `dazzled' by the presence of words that have a grounded referent in the image (e.g., \textit{woman}, \textit{girl}, or \textit{lady} in some of the unrelated descriptions), that could lead it to assign some similarity even when the sentence is completely out of place. Conversely, the absence of words, and particularly nouns, with a clear grounded referent in the \textsc{Full} description would be considered by the model as an indicator of misalignment.
This could be a result of the model training data and learning objective. On the one hand, the 
$\langle$image, text$\rangle$ pairs scraped from the web may be poorly representative of language uses in real communicative contexts, where semantic underspecification is ubiquitous. On the other hand, the \textit{contrastive} learning objective being employed may be too aggressive with texts that do not conform to those typically seen in training. In both cases, the similarity assigned to an underspecified description would be lower than the 
(possibly small) 
similarity assigned to an unrelated sentence with one or a few matching elements.

\paragraph{Moving forward}
Taken together, the results of the two PoCs show that CLIP struggles with semantically underspecified language. This limitation must be taken into consideration if we want to use this and similar systems to model real communicative scenarios or use them in applications that interact with human users---which is not the case for most of the tasks these models are trained and tested on.
Indeed, these models may fail to retrieve an image if the language query used does not conform to the standard type of descriptions seen in training. Or, they could misunderstand inclusive uses of certain pronouns (e.g., \textit{they}), and exhibit unwanted overspecification biases when producing an image description or referring utterance. 
We argue that our community, if it aims at developing language technology that can successfully and efficiently communicate with human users, should be aware of semantic underspecification and take steps toward making our models master it properly.

In the next section, we discuss how this is relevant to a range of studies exploring multimodal tasks in communicative settings.

\section{Communicative Approaches to Multimodal Tasks}\label{sec:communication}

Mastering semantic underspecification is relevant to  
a wide range of studies that take a \textit{communicative} or pragmatic approach to multimodal tasks. Below, we focus on a select sample of them\footnote{For a recent and exhaustive survey of pragmatic work in multimodal NLP, see~\citet{fried2022pragmatics}.} and discuss how they might benefit from a full mastery of the phenomenon  investigated in the paper.

\paragraph{Image captioning with a communicative goal} Standard image captioning\footnote{Note that, throughout this paper, we consistently avoid the term \textit{captions} and instead use \textit{descriptions}. We believe this terminology better reflects the fact that IC aims at generating faithful descriptions---and not captions---of images~\cite[see also a similar argument in][]{kreiss2021concadia}.} consists in generating a description that is as close as possible to the content of the image. Typically, the task is not tied to a real communicative goal: image descriptions are provided by crowdworkers who are asked to mention all the important aspects of an image~\cite{hodosh2013framing,lin2014microsoft,young2014image},\footnote{One exception is represented by VizWizCap~\cite{gurari2020captioning}, a dataset of image descriptions collected with the explicit aim to be informative to visually impaired people.} and models are trained and evaluated to closely approximate those descriptions~\cite{bernardi2016automatic}.
To make the task more pragmatically valid, some work proposed a discriminative version of it where models need to generate a description for an image that is pragmatically informative, i.e., that is good for the image in the context of other distractor images~\cite{andreas2016reasoning,vedantam2017context,cohn2018pragmatically,nie2020pragmatic}. The recent Concadia dataset~\cite{kreiss2021concadia}, in contrast, considers images in isolation and focuses on the communication needs of image captioning. In particular, 
it distinguishes between \textit{descriptions}, useful to describe an image to someone who does not have access to the image, and \textit{captions}, that instead complement the information of an image that is available to both interlocutors.\footnote{Captions of images in the context of news articles are a prototypical example~\cite{hollink2016corpus,biten2019good}.} 

Within both lines of research, it is to be expected that underspecified language comes into play. For example, neither the gender nor the number of people in an image may be needed to pragmatically distinguish it from other images; or, a caption complementing an image (and an accompanying text) may leave underspecified much of the information that one can get from elsewhere.\footnote{It is worth mentioning that also the visual features of the scene to be described could play a non-negligible role in this sense, as recently shown by~\citet{berger-etal-2023-large}.} As such, these tasks would greatly benefit from having models opportunely dealing with this language phenomenon. In support of this---and in line with the results of our PoCs above---recent work~\cite{kreiss2022context} showed that SotA models like CLIP are unable to account for the degree of \textit{usefulness} of an image description, but only for its alignment. More generally, focusing on semantic underspecification of visually grounded language would be relevant to studies investigating the range of relations that texts entertain with images, including communicative goals and information needs~\cite{kruk2019integrating,alikhani-etal-2020-cross}. Moreover, it would inform the \textit{specular} task of image-to-text generation, as recently claimed by~\citet{hutchinson2022underspecification}.

\paragraph{Goal-oriented visual question answering} Standard visual question answering datasets~\cite{antol2015vqa} have been collected by asking crowdworkers to provide questions and answers for research purposes. In contrast, the VizWiz dataset~\cite{gurari2018vizwiz} includes questions that were asked by visually-impaired people to obtain information about visual scenes. As such, these questions are motivated by a real communicative goal and exhibit very different linguistic features compared to the questions and answers in standard datasets. For example, the questions are more ambiguous or underspecified, and the answers by the respondents are more diverse and subjective~\cite{yang2018visual,bhattacharya2019does,jolly2021ease}. We propose that models that are equipped for semantically underspecified language should both better understand the question in relation to an image~\cite[something that current SotA models struggle with, see][]{chen2022grounding} and better leverage the diversity and sparseness of the answers. 

Similarly, these models may better integrate the complementary information conveyed by language and vision in, e.g., \textsc{bd2bb}~\cite{pezzelle2020different}, a version of the visual question answering task where the correct answer (an action) results from the combination of a context (an image) and a fully \textit{ungrounded} intention (a text); or, in other datasets that require abductive reasoning~\cite{hessel2022abduction}. Finally, models that master semantic underspecification are expected to also deal better with related phenomena found in visual question answering, such as ambiguity and vagueness, highlighted in~\citet{bernardi2021linguistic}.

\paragraph{Object naming and referring expressions}
Multimodal models should be robust to variation in object naming. For example, they should not consider as an error the use of  the noun \textit{artisan} to refer to the person in Figure~\ref{fig:errors}, even if another noun, e.g., \textit{person}, was perhaps used more frequently. At the same time, the degree of semantic specification of a naming expression should be accounted for, which would be needed to replicate patterns on human naming variation, as the ones reported by~\citet{silberer-etal-2020-humans} and~\citet{gualdoni2022woman}.

Naming variation is also observed in more complex visually grounded reference games, where the task is to produce a referring expression that is pragmatically informative, i.e., that allows a listener to pick the target object (image). This task is the ideal benchmark for testing how various pragmatic frameworks, such as the Rational Speech Acts~\cite[RSA;][]{frank2012predicting,goodman2016pragmatic}, can model the reference to, e.g., colors~\cite{monroe2017colors} in artificial settings. 

Turning to naturalistic scenarios, recent work used CLIP
to quantify the properties of human referring expressions. The model was shown to capture the degree of \textit{discriminativeness} of a referring expression over a set of images, though it assigned lower alignment scores (computed without taking into account the broader visual context) to progressively more compact utterances~\cite{takmaz-etal-2022-less}. Our PoCs above showed that this model conflates the semantic validity of a description with its degree of over or underspecification. However, distinguishing between the two is crucial, e.g., to assess that the expressions \textit{the guy with the tattoos} and \textit{the tattoo guy} are semantically equally valid, with the latter being just more underspecified (the semantic relation tying the constituents of the compound has to be inferred from the image). This can lead to models that are capable of reproducing human language patterns in certain communicative scenarios~\cite[e.g., the shortening and compression of referring utterances over an interaction, see][]{takmaz2020refer} without explicit supervision.

\paragraph{Visually-grounded goal-oriented dialogue}
All the abilities mentioned above are relevant to the development of dialogue systems that can entertain a goal-oriented conversation with human users. Examples of visually grounded goal-oriented dialogue encompass reference tasks where either \textit{yes/no} questions~\cite{de2017guesswhat} or free-form, open-ended dialogue utterances~\cite{udagawa2019natural,ilinykh2019meet,haber2019photobook} are allowed to achieve a common goal, e.g., figuring out what object is being talked about or is in common between the two players. Most of these studies use datasets of interactions between human speakers to train systems that can learn to have a successful dialogue while reproducing similar linguistic and pragmatic patterns. In a few notable exceptions~\cite{liu2018dialogue,hawkins-etal-2020-continual}, these systems entertain an actual interaction with human users and go through a process of continual learning that leverages that \textit{online} data.
Given the communicative nature of the task, semantic underspecification is likely to be an important feature of the language used. In particular, it appears to deserve special attention when the goals involve giving and receiving visually grounded instructions~\cite[here, it is indeed one of the dimensions considered when analyzing models' results; see][]{10.1162/tacl_a_00428}. Once again, models must be capable of dealing with semantic underspecification to communicate successfully and efficiently.

In the next section, we outline a few research directions and provide examples of concrete steps that can guide work aimed at achieving this goal.

\section{Research Directions}\label{sec:directions}

\subsection{Definitions and Operationalizations} 
As discussed in Section~\ref{sec:intro}, semantic underspecification can be generally defined as the lack, in a linguistic signal, of part of the semantic information required to understand the message, which is typically obtained from other linguistic and non-linguistic sources. To tackle the problem at a computational level, it is important to formally define and operationalize the phenomenon. For example,
by identifying which linguistic phenomena, words, or classes of words are considered by the linguistic theory as instances of semantic underspecification and under which circumstances (top-down approach). Or, by means of a data-driven measure, such as the applicability of a text to a more or less large number of visual contexts (bottom-up approach). 
In either case, computational methods can be used to refine or validate such definition~\cite[this is the approach used, for example, by a recent work testing the Uniform Information Density theory using language models;][]{giulianelli-etal-2021-information}. Moreover, computational methods may be used to distinguish between instances of underspecification that are \textit{deliberate} (e.g., using the pronoun \textit{they} to refer to an individual) from those that may depend on contextual or situational aspects (e.g., not having access to some information or not mentioning something that is socially and culturally obvious).

\subsection{Datasets and Annotations} 
Novel datasets or \textit{ad hoc} annotations of existing resources can be collected to study underspecified language. 
These datasets can encompass the standard multimodal tasks (image captioning, visual question answering, etc.) and therefore be used as evaluation benchmarks to test existing models; or, new tasks can be proposed, including the prediction of an underspecification score, the paraphrasing or explanation of an underspecifed sentence (or, \textit{vice versa}, the de-overspecification of a sentence), and so on. 
Moreover, annotations may be collected at the sample and dataset level to investigate, for example, whether overspecified and underspecified image descriptions or referring utterances are equally good, informative, or inclusive\footnote{These directions may also be relevant to the line of work exploring how to minimize biases and misrepresentations when describing images~\cite[e.g.,][]{bennett2021s}.} according to human speakers, how many and which non-linguistic cues are needed to understand them, which visual and communicative contexts elicit more underspecified language, and so on.

\subsection{Model Training and Testing}
Operationalizing and annotating semantic underspecification can be useful, in turn, for training and testing purposes. As for the former, sampling cases from a dataset with a varying degree of semantic underspecification can be helpful for training or fine-tuning models to make them more robust to any language. As for the latter, benchmarking a model with underspecified language can shed light on its generalization abilities and applicability to truly communicative scenarios. Moreover, a measure of a sample's semantic underspecification could be used as an additional learning signal for the training of foundational, task-agnostic multimodal models. Indeed, such a measure may indicate the extent to which language and vision convey redundant or complementary information, the relative importance of each modality, and the relation between the correctness and \textit{self-sufficiency} of a sample. Finally, it may be interesting to leverage the degree of semantic underspecification as a dimension to which NLG models can adapt, e.g., to generate text that is more or less specified depending on the context, the interlocutor's needs or style, and the communicative goal of the linguistic interaction.

\section{Conclusion}

In this position paper, we argued that the NLP community must deal with semantic underspecification, that is, 
the possibility for a linguistic signal to convey only part of the information needed to understand a message. This is a ubiquitous phenomenon in human communication, that speakers deal with by quickly and  effortlessly integrating non-linguistic information, e.g., from the surrounding visual context. We argued that research in multimodal NLP combining language and vision is ready to take on this challenge, 
given that 
SotA models that achieve unprecedented performance on a range of downstream tasks (image captioning, visual question answering, etc.) appear to struggle with it. We indicated several directions and concrete steps toward achieving this goal and discussed tasks and applications that would benefit from a full mastery of semantic underspecification.

On a technical level, our paper highlights the need to improve SotA models by making them robust to scenarios that may be different from those seen in training. In our case, CLIP suffers with sentences that resemble the language used in real communicative contexts, which poses a problem if we were to use it for modeling communicative tasks or embed it in user-facing applications. This general weakness of SotA models has been recently illustrated by~\citet{thrush2022winoground}. Using  WinoGround, a dataset of carefully designed $\langle$image, description$\rangle$ pairs testing compositionality abilities, the authors reported chance-level performance for all the Transformer-based multimodal models they tested---including CLIP. A careful analysis of the samples by~\citet{diwan2022winoground} revealed that the difficulties of the dataset go beyond dealing with compositionality, and include ambiguity aspects, reasoning abilities, and so on. In any case, these findings are informative of the flaws of the models and provide useful indications on which directions to take for improving them.

On a theoretical level, the ideas presented in our paper are consonant with a recent line of thought that advocates approaches that are aware of communicative and pragmatic aspects in language understanding and generation~\cite{andreas2022language,fried2022pragmatics,giulianelli2022towards,schlangen-2022-norm}. We believe this is an exciting direction, and support a collaborative effort aimed at developing systems that can use language 
with a communicative goal.

\section*{Limitations}

Semantic underspecification has been extensively studied in semantics, pragmatics, psycholinguistics, communication sciences, and cognitive sciences. In this position paper, we review this literature only superficially, although we are aware that
a generalized and exhaustive understanding of the phenomenon necessarily requires knowledge of this previous work. We encourage the scholars working on this topic to embrace its complexity and depth. 

The paper focuses on approaches, tasks, and models within multimodal NLP. As such, it almost completely neglects a discussion of semantic underspecification within text-only NLP. However, we are aware of the growing interest in the community at large for frameworks that propose and evaluate models in pragmatic or communicative contexts~\cite[][]{pragmaticLM,andreas2022language,hu_fine-grained_2023}, and that some of the directions and steps that we propose could apply to text-only models~\cite[see, e.g., the recent, relevant work on large language models and ambiguity by][]{liu2023we}.

The two proofs of concept we report in the paper consider a rather narrow set of semantic underspecification phenomena, which may not be entirely representative. Moreover, the manual annotation that we perform, though consistent, does not adhere to any strict guidelines, and borderline cases are entrusted to the linguistic competence of the annotator. Finally, and more in general, these proofs of concepts are mostly intended to serve as a basis for the discussion and as an indication of patterns and trends. Therefore, future work should further and more thoroughly investigate this issue.

\section*{Acknowledgements}
This paper owes much to the constant and passionate dialogue with the members of the Dialogue Modelling Group at the ILLC, particularly Mario Giulianelli, Ece Takmaz, and Alberto Testoni. A special thanks goes to Raquel Fernández for her valuable comments on a draft of the article.

\bibliography{anthology,custom}
\bibliographystyle{acl_natbib}

\end{document}